\documentclass{article}
\usepackage{times}
\usepackage{subfigure}
\usepackage{graphicx}
\usepackage{amssymb}
\usepackage{booktabs}
\usepackage{multirow}
\usepackage{multicol}
\usepackage{arydshln}
\usepackage{booktabs}  
\usepackage{threeparttable}
\usepackage{algorithm}
\usepackage{algorithmic}
\usepackage{setspace}
\usepackage{spconf,amsmath,epsfig}
\usepackage{bm}
\usepackage{fancyhdr}

\let\OLDthebibliography\thebibliography
\renewcommand\thebibliography[1]{
	\OLDthebibliography{#1}
	\setlength{\parskip}{0pt}
	\setlength{\itemsep}{0pt plus 0.3ex}
}

\begin{document}\sloppy
	
	\def\x{{\mathbf x}}
	\def\L{{\cal L}}

	\title{LRNNet: A Light-Weighted Network with Efficient Reduced Non-Local Operation for Real-Time Semantic Segmentation}
	%
	%
	\name{Weihao Jiang\qquad  Zhaozhi Xie\qquad  Yaoyi Li\qquad Chang Liu\qquad Hongtao Lu\thanks{$^{\dagger}$Corresponding author}$^{\dagger}$}
	\address{Department of Computer Science and Engineering, MoE\\ 
Key Lab of Artiﬁcial Intelligence, AI Institute, Shanghai Jiao Tong University, China\\
\{jiangweihao, xiezhzh, dsamuel, isonomialiu, htlu \}@sjtu.edu.cn}

	\maketitle
	
\thispagestyle{fancy}
\fancyhead{}
\lhead{}
\lfoot{}
\cfoot{}
\rfoot{978-1-7281-1485-9/20/\$31.00 ©2020 IEEE}

	\begin{abstract}
		The recent development of light-weighted neural networks has promoted the applications of deep learning under resource constraints and mobile applications. Many of these applications need to perform a real-time and efficient prediction for semantic segmentation with a light-weighted network. This paper introduces a light-weighted network with an efficient reduced non-local module (LRNNet) for efficient and real-time semantic segmentation. We proposed a factorized convolutional block in ResNet-Style encoder to achieve more light-weighted, efficient and powerful feature extraction. Meanwhile, our proposed reduced non-local module utilizes spatial regional dominant singular vectors to achieve reduced and more representative non-local feature integration with much lower computation and memory cost. Experiments demonstrate our superior trade-off among light-weight, speed, computation and accuracy. Without additional processing and pre-training, LRNNet achieves 72.2\% mIoU on Cityscapes test dataset only using the fine annotation data for training with only 0.68M parameters and with 71 FPS on a GTX 1080Ti card.
	\end{abstract}
	\section{Introduction}
	\label{sec:intro}
	
	Semantic segmentation can be viewed as a task of pixel-wise classification, which assigns a specific pre-defined category to each pixel in an image. The task has many potential applications in autonomous driving or image editing and so on. Although many works \cite{fcn,DANet,CCNet} have made great progress in the accuracy of image semantic segmentation tasks, their network size, inference speed, computation and memory cost limit their practical applications. Therefore, it's essential to develop the light-weighted, efficient and real-time methods for semantic segmentation.
	\begin{figure}[t]
		\centering
		\begin{tabular}{@{}c@{}} 
			\includegraphics[width=.44\textwidth,height=.20\textwidth]{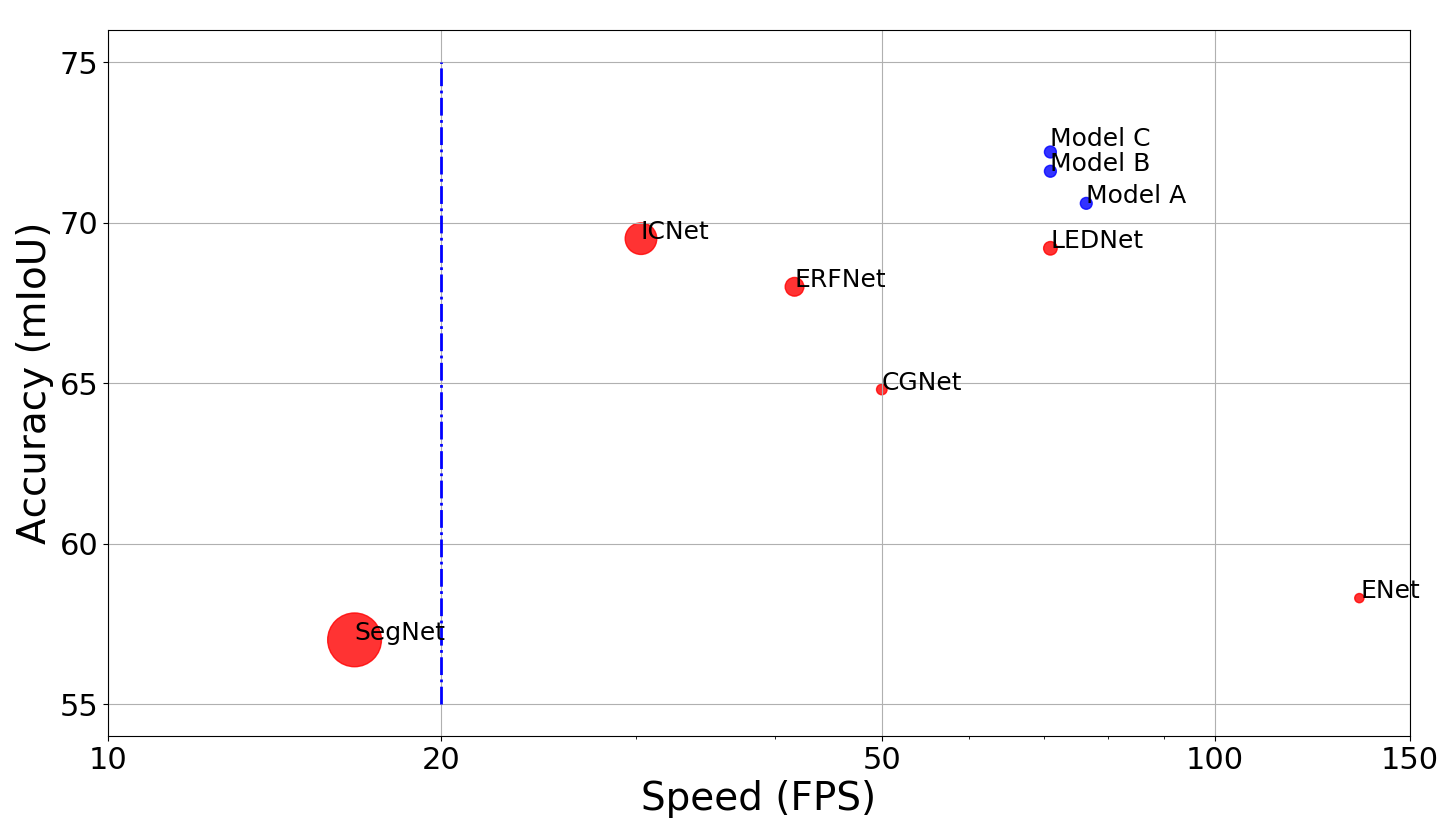} 
		\end{tabular}
		\caption {Inference speed, accuracy and learnable network parameters on Cityscapes \cite{cityscapes} test set. The smaller bubble means fewer parameters. We compare LRNNet with methods implemented in open-source deep-learning frameworks, such as Pytorch \cite{pytorch} and Caffe, including ICNet \cite{ICNet}, CGNet \cite{CGNet}, ERFNet \cite{ERFNet}, SegNet \cite{Segnet}, ENet \cite{ENet} and LEDNet \cite{wang2019lednet}.}. 
		\label {fig1}
	\end{figure}
	Among these properties, light-weight could be the most essential one, because  using a smaller scale network can  lead to faster speed and more efficient in computation or memory cost easier. A less learnable network parameter means that the network has less redundancy and network structure makes its parameters more effective. Practically, a smaller model size is more favorable for cellphone apps. Designing blocks with proper factorized convolution to construct an effective light-weighted network could be a more scalable way and could be easier to balance among accuracy, network size, speed, and efficiency. In this way, many works \cite{ENet,BiSeNet,ERFNet,wang2019lednet,li2019dfanet} achieve promising results and show their potential for many potential applications. But those works \cite{ERFNet,wang2019lednet} do not balance factorized convolution and long-range features in a proper way.
	
	The recent study \cite{Non-local} shows the powerful potential of attention mechanism in computer vision. Non-local methods are employed to model long-range dependencies in semantic segmentation \cite{DANet}. However, modeling relationships between every position could be rather heavy in computation and memory cost. Some works try to develop factorized \cite{CCNet} or reduced \cite{Asymmetric} non-local methods to make it more efficient. Since efficient non-local or positional attention is not developed enough for light-weighted and efficient semantic segmentation, our approach tries to develop a powerful reduced non-local method to model long-range dependencies and global feature selection efficiently.
	
	In our work, we develop a light-weighted factorized convolution block (FCB) (Fig. \ref{fig:FCB}) to build a feature extraction network (encoder), which deals with long-range and short-range features with proper factorized convolution respectively, and we proposed a powerful reduced non-local module with regional singular vectors to model long-range dependencies and global feature selection for the features from encoder to enhance the segmentation results. Contributions are summarized as follows:
	\begin{itemize}
		\item We proposed a factorized convolution block (FCB) to build a very light-weighted, powerful and efficient feature extraction network by dealing with long-range and short-range features in a more proper way.
		\item The proposed efficient reduced non-local module (SVN) utilizes regional singular vectors to produced more reduced and representative features to model long-range dependencies and global feature selection.
		\item All experiments show the state-of-the-art trade-off in terms of parameter size, speed, computation and accuracy of our LRNNet on Cityscapes \cite{cityscapes} and Camvid \cite{Camvid} datasets.
	\end{itemize}
	
	\begin{figure*}
		\centering
		\subfigure[Our Encoder Network]{\label{Arc:a}\includegraphics[width=.45\textwidth,height=.18\textwidth]{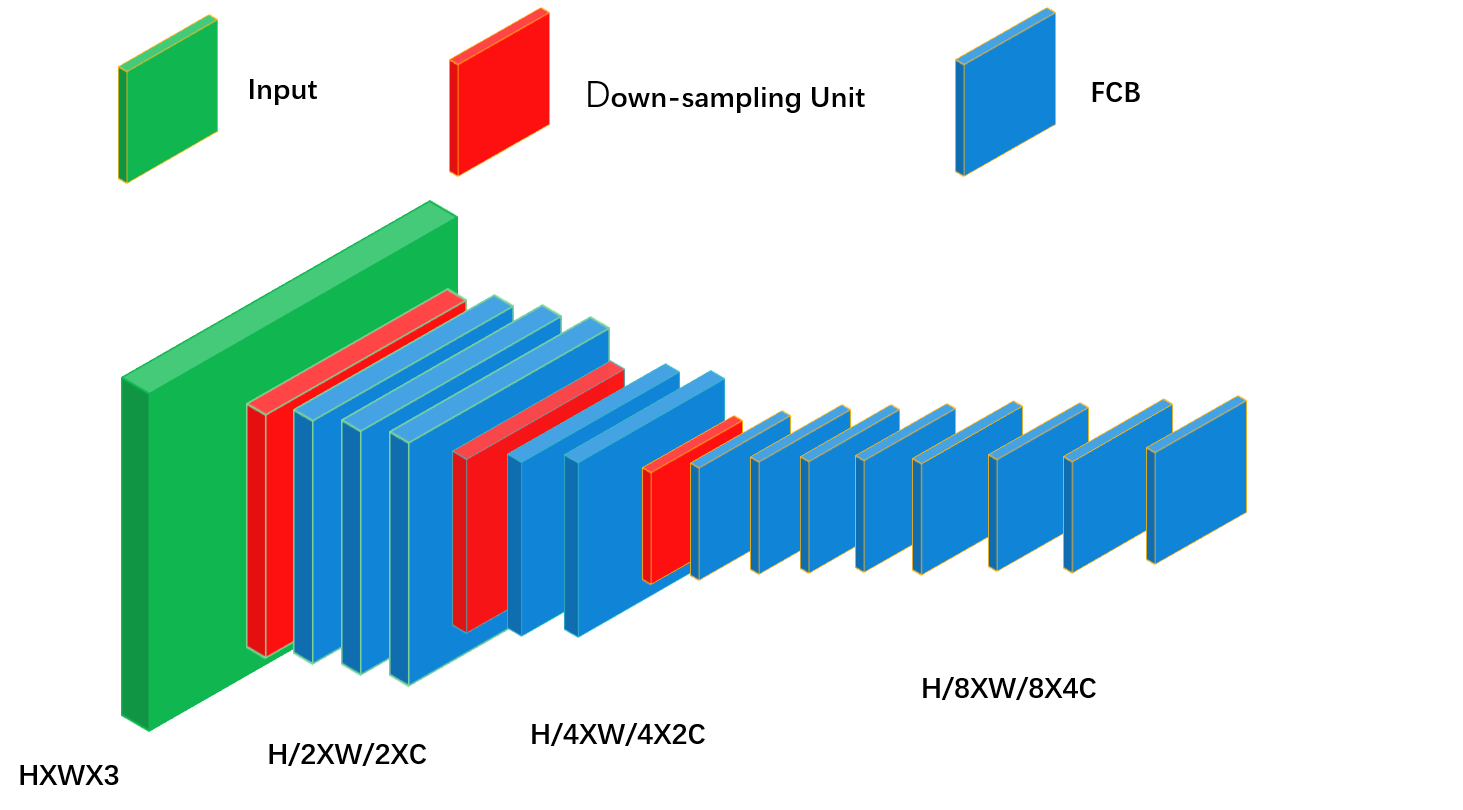}}
		\subfigure[Our Decoder (classifier) with SVN Module]{\label{Arc:b}\includegraphics[width=.45\textwidth,height=.18\textwidth]{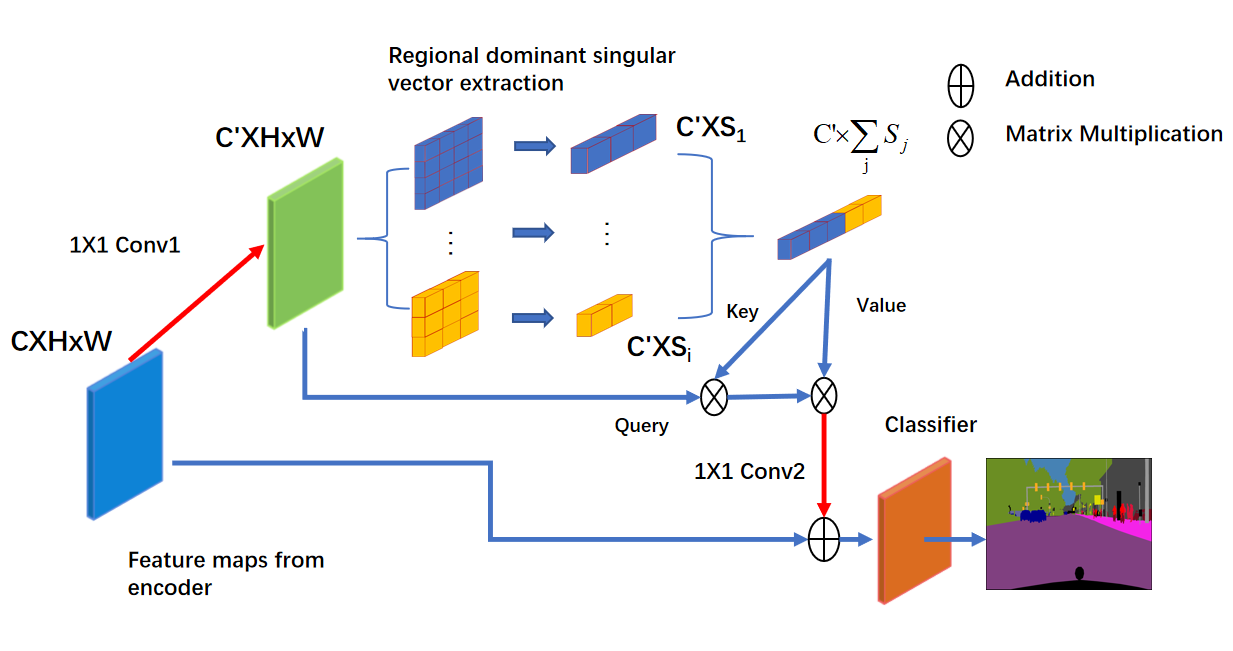}}
		\caption{Overview of our LRNNet: (a) shows our light-weighted encoder which is constructed by our factorized convolution block (FCB) in a three-stages ResNet-style. (b) shows our decoder with SVN module. The upper branch performs our non-local operation with regional dominant singular vectors in different scales. The red arrows represent $1\times 1$ convolutions adjusting the channel size to forming an bottleneck structure. The classifier consists a $3\times 3$ convolution followed by a $1\times 1$ convolution.}  
		\label{Arc}
	\end{figure*}
	\section{Related Work}
	
	\textbf{Light-weighted and Real-time Segmentation.} Real-time semantic segmentation approaches aim to generate high-quality prediction in limited time, which is usually performed under resource constraints or mobile applications. Light-weight models save storage space and potentially have lower computation and faster speed. Therefore, developing light-weight segmentation is a potential way to get a good trade-off for real-time semantic segmentation \cite{CGNet,wang2019lednet,ERFNet}. Our model follows the light-weight style to achieve real-time segmentation.
	
	\textbf{Factorized Convolution.} Standard convolution adopts a 2D convolution kernel to form a full connection between input and output channels, which learns local relation and channel interaction. However, this may suffer from the large parameter size and redundancy for real-time tasks under resource constraints. Xception \cite{Xception} and MobileNet \cite{MobileNetv1} adopt depthwise separable convolution, which consists of a depthwise convolution followed by a pointwise convolution. Depthwise convolution learns local relation in every channel and pointwise convolution learns the interaction between channels to reduce parameters and computation. ShuffleNet \cite{ShuffleNetv1} adopts a split-shuffle strategy to reduce parameters and computation. In this strategy, standard convolution is split into some groups of channels and a channel shuffle operation helps the information flows between groups. Factorizing the 2D convolution kernel into a combination of two 1D convolution kernels is another way to reduce parameter size and computation cost. Many light-weight approaches \cite{wang2019lednet,ERFNet} take this way and get promising performances. In this paper, our convolution factorization block (FCB) utilizes these strategies to build a light-weighted, efficient and powerful structure.
	
	\textbf{Attention Models.} Attention modules model long-range dependencies and have been applied in many computer vision tasks. Position attention and channel attention are two important mechanisms. Channel attention modules are widely applied in semantic segmentation \cite{li2019dfanet} including some light-weighted approaches \cite{CGNet,li2019dfanet}. Position attention or non-local methods have a higher computational complexity. Although some works \cite{CCNet,Interlaced,Asymmetric} try to develop more efficient non-local methods, position attention or non-local methods are rarely explored in light-weighted semantic segmentation. 
	
	\section{Methodology}
	We introduce the preliminary related to our SVN module in section 3.1, network architecture in section 3.2, the proposed FCB unit in section 3.3 and SVN module in section 3.4.
	\subsection{Preliminary}
	Before introducing the proposed method, we first introduce the singular value decomposition and non-local method, which are related to our SVN module in section 3.4.
	
	\textbf{Singular Value Decomposition and Approximation.} Given a real matrix $A = (a_{ij})\in R^{m\times n} (m\geq n)$, with real numbers $\sigma _1\geq \sigma _2\geq ...\geq \sigma _r > 0$, there exist two orthogonal matrices $U \in R^{m\times m}$ and $V \in R^{n\times n}$, satisfying Equation \ref{svd},
	\begin{equation}
	A = UDV^{T}=\sum_{i=1}^{r}{\sigma _i u_i v_i^{T}} 
	\label{svd}
	\end{equation}
	where $D = diag\{ \sigma _1, \sigma _2, ..., \sigma _r, 0, ..., 0 \} _{m\times n}$, $U =\{ u _1, u _2, ...,u _m \}$ and $V =\{ v _1, v _2, ...,v _n \}$.
	If we choose $K\leq r$, we can get
	\begin{equation}
	A\approx \sum_{i=1}^{K}{\sigma _i u_i v_i^{T}}=\hat{A}
	\end{equation}
	where $\bm{\hat{A}}$ approximates the original $\bm{A}$, because the larger singular values and their singular vectors keep most of the information of $\bm{A}$. The corresponding singular vectors of the larger singular value contain more information of the matrix, especially the dominant singular vectors. We can calculate the dominant singular vectors by power iteration Algorithm \ref{PI} efficiently. Based on Equation \ref{svd}, rotating columns of $\bm{A}$ does not change $\bm{U}$ and $\bm{u_i}$ and their singular values.
	\begin{algorithm}[!h]
		\caption{Power Iteration$(A,T)$}
		\label{PI}
		\begin{algorithmic}[1]
			\STATE Input: $A\in R^{m\times n}$,  iterations: T and init $u\in R^{m}$
			\FOR{$i\  in\  T$}
			\STATE $u := \frac{u}{||u||_2}$, $v := A^{T}u$
			\STATE $v := \frac{v}{||v||_2}$, $u := Av^{T}$
			\ENDFOR
			\STATE return $u := \frac{u}{||u||_2}$
		\end{algorithmic}
	\end{algorithm}
	
	\textbf{Non-local Module.} Non-local module \cite{Non-local} models global feature relationships. We illustrate it in the form of Query-Key-Value. It can be formulated as:
	\begin{equation}
	O_i =\frac{1}{C(v_j)}\sum_{j=1}{Sim(q_i,k_j)v_j}
	\label{non}
	\end{equation}	
	where $\bm{q_i }\in \bm{Q}, \bm{k_j}\in \bm{K}, \bm{v_j}\in \bm{V}$, $\bm{q_i}\in R^{C_1}$ is a Query, $\bm{O_i}$ is the corresponding output of $\bm{q_i}$, $\bm{k_j}\in R^{C_1}$, is a Key, $\bm{v_i}\in R^{C_2}$, is a Value, $Sim(\cdot , \cdot )$ is the measure of similarity between $\bm{k_i}$ and $\bm{q_i}$, and $C(x)$ is a normalization function, $\bm{Q}\in R^{C_1\times N_1}$, $\bm{K}\in R^{C_1\times N_2}$ and $\bm{V}\in R^{C_2\times N_2}$ are the collections of the Queries, the Keys and the Values, respectively. And a smaller $N_2$ means less computation.
	
	\subsection{Overview of the Network Architecture}
	In this section, we introduce our network architecture. Our LRNNet consists of a feature extraction network constructed by our proposed factorized convolution block (Fig.  \ref{fig:FCB:c}) and a pixel classifier enhanced by our SVN module (Fig. \ref{Arc}).
	
	We form our encoder in a three-stages ResNet-style ( Fig. \ref{Arc:a}). We adopt the same transition between stages as ENet \cite{ENet} using a downsampling unit.  The core components are our FCB units, which provide light-weighted and efficient feature extraction. For better comparison of other light-weight factorized convolution block, we adopt the same dilation series of FCB in encoder as LEDNet \cite{wang2019lednet} after the last downsampling unit (details in supplemental material). Our decoder ( Fig. \ref{Arc:b}) is a pixel-wise classifier enhanced by our SVN module. 
	\begin{figure*}[t]
		\centering
		\renewcommand\tabcolsep{0.6 pt}
		\begin{tabular}{@{}cccccc@{}}
			\includegraphics[width=.14\textwidth,height=.06\textwidth]{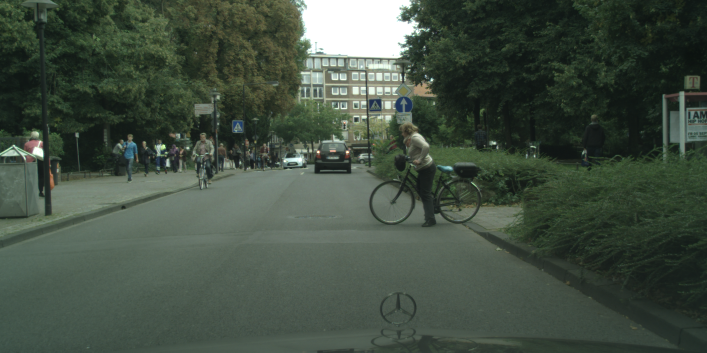} &
			\includegraphics[width=.14\textwidth,height=.06\textwidth]{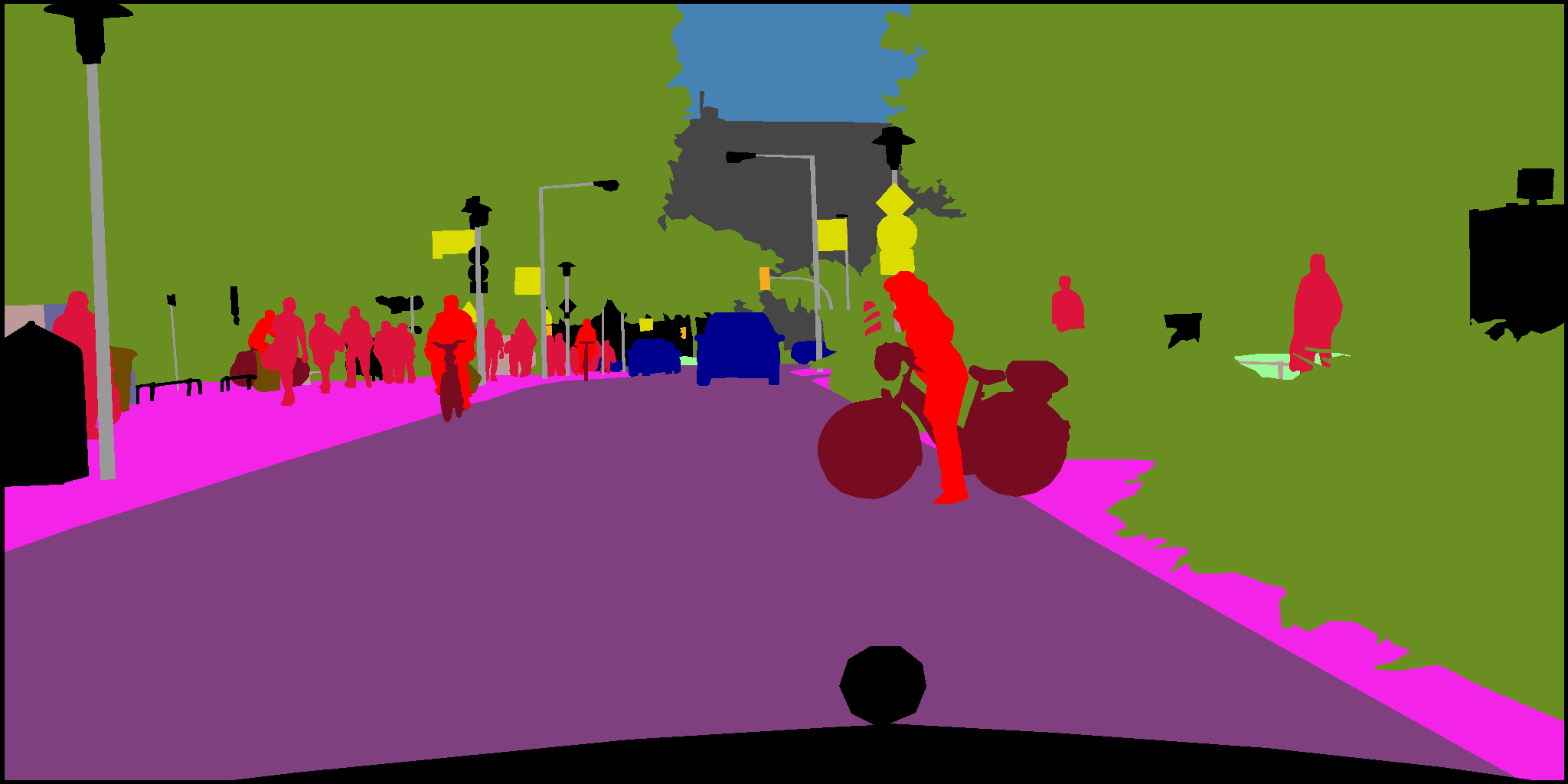} &
			\includegraphics[width=.14\textwidth,height=.06\textwidth]{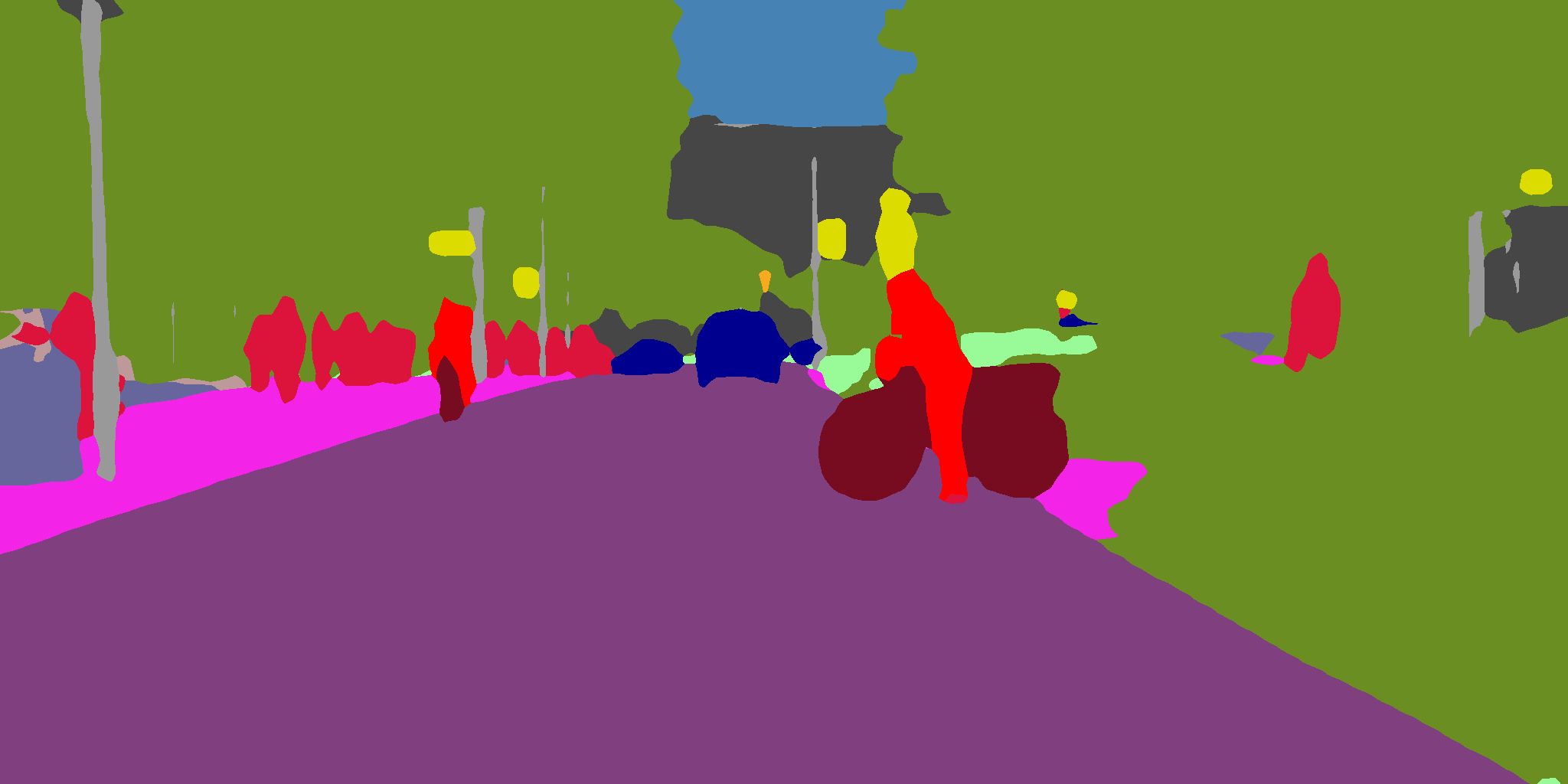} &
			\includegraphics[width=.14\textwidth,height=.06\textwidth]{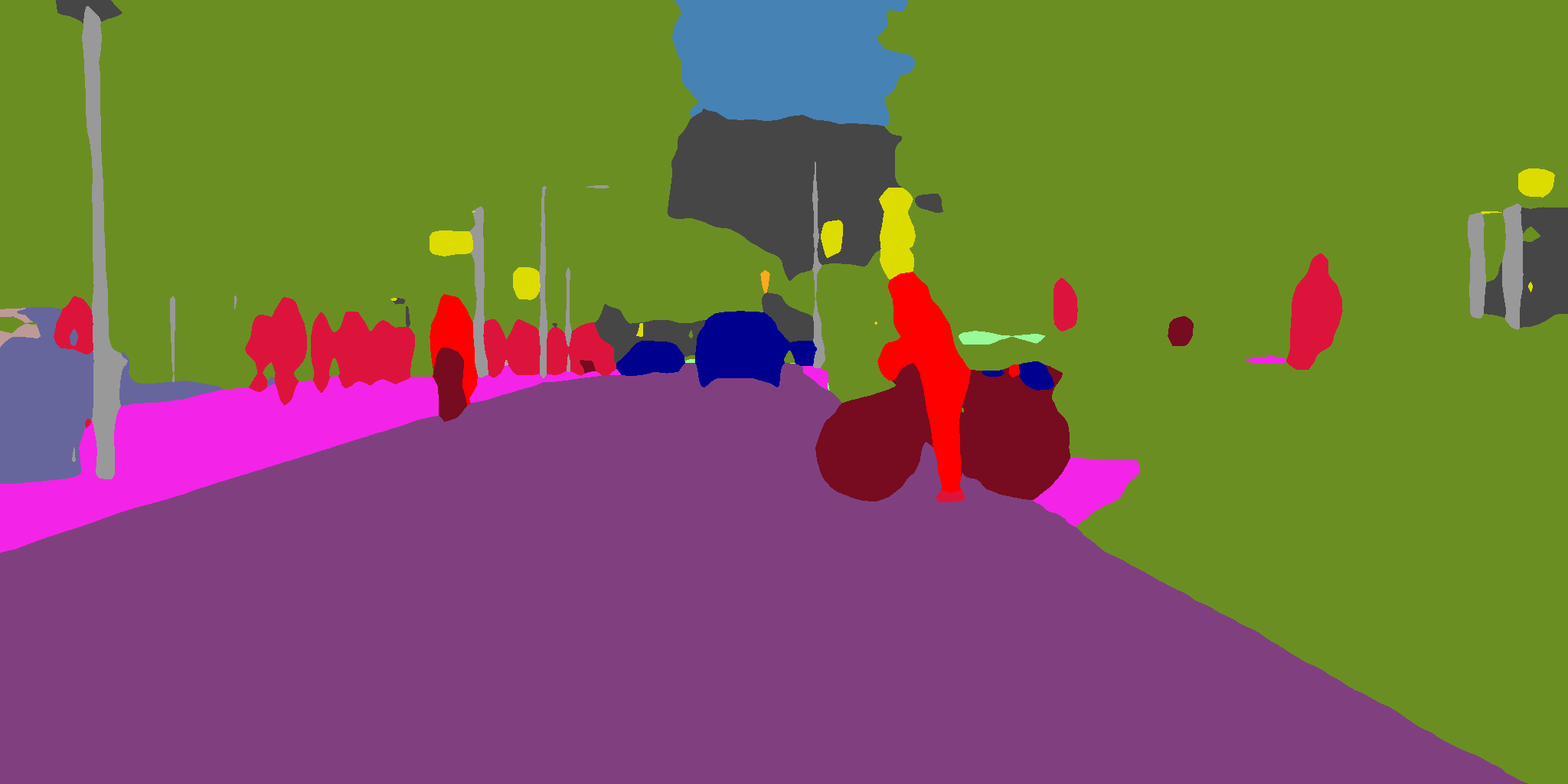} &
			\includegraphics[width=.14\textwidth,height=.06\textwidth]{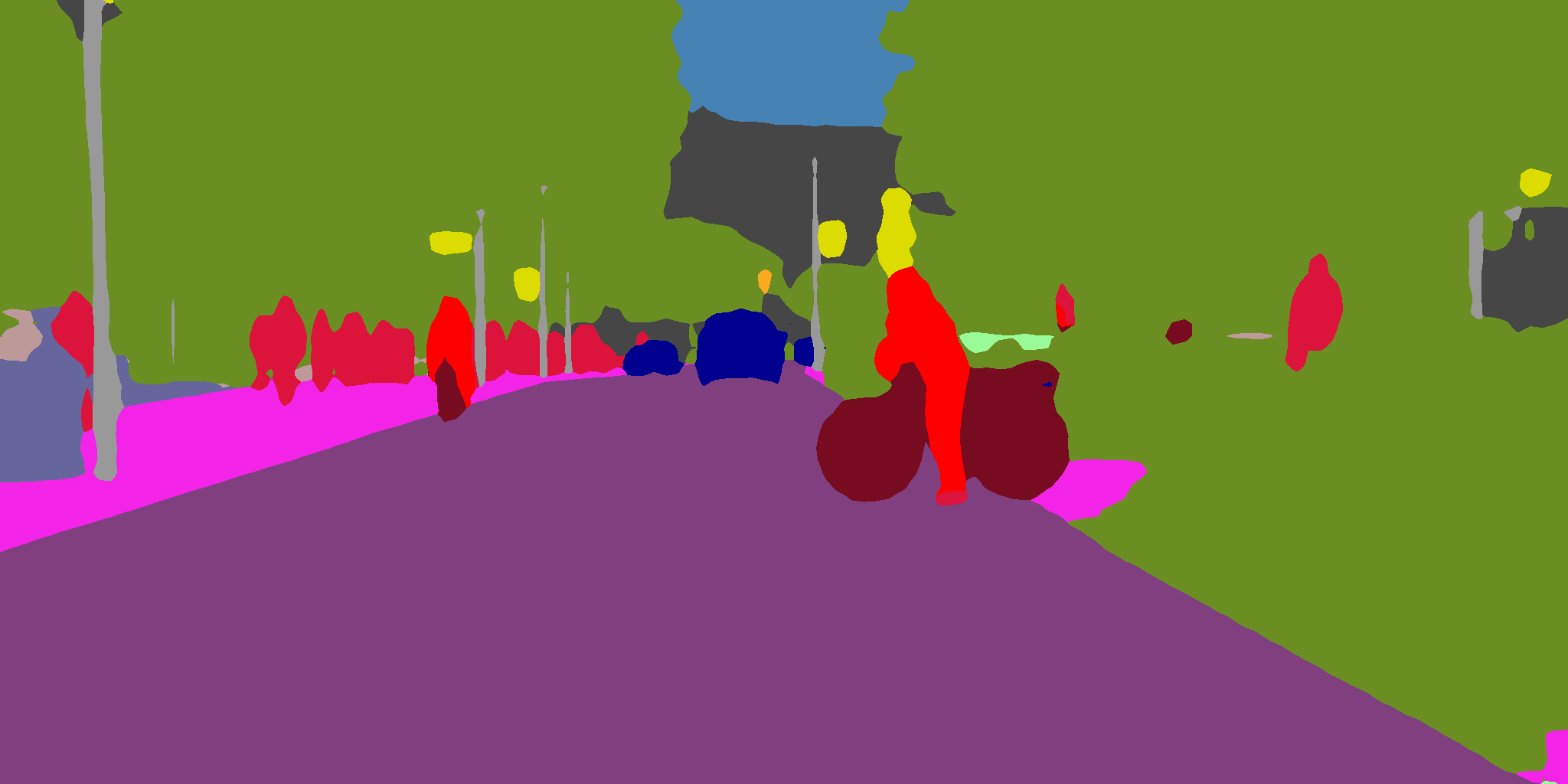} &
			\includegraphics[width=.14\textwidth,height=.06\textwidth]{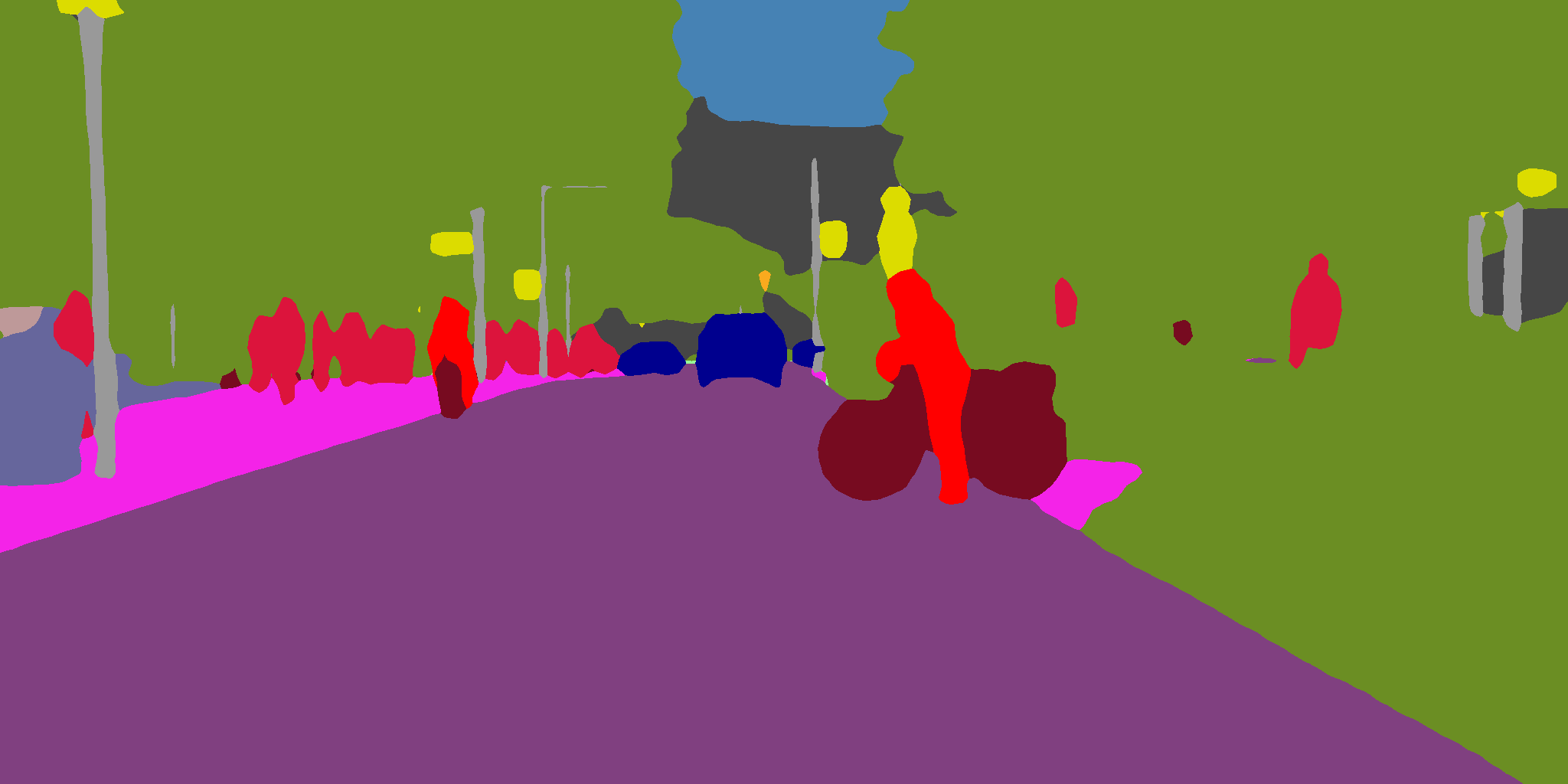}   \\
			\includegraphics[width=.14\textwidth,height=.06\textwidth]{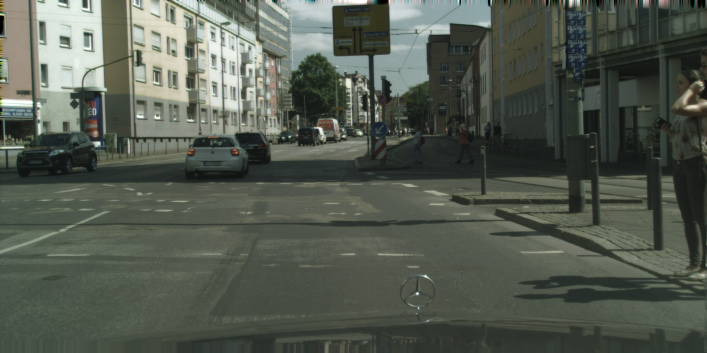} &
			\includegraphics[width=.14\textwidth,height=.06\textwidth]{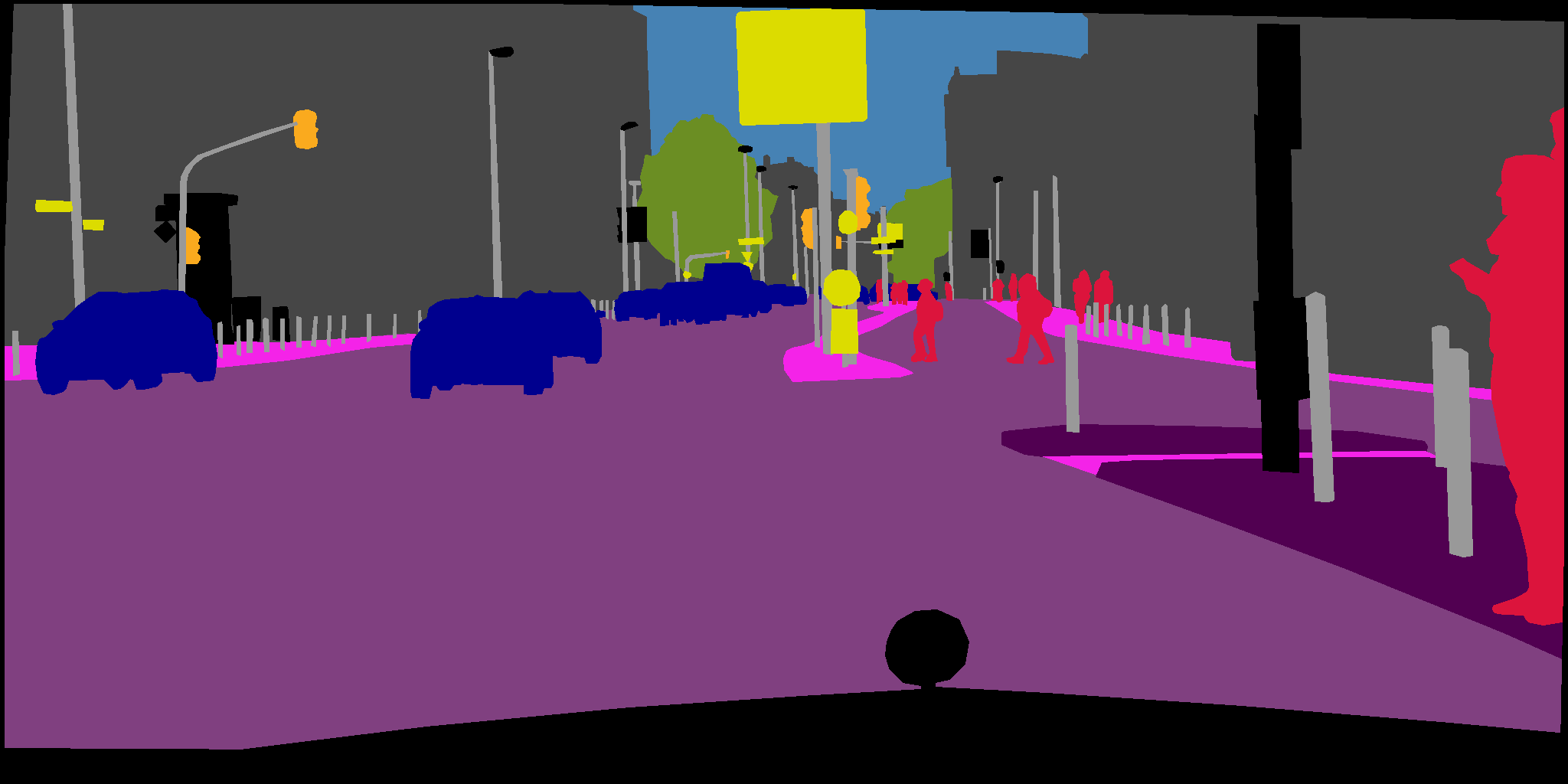} &
			\includegraphics[width=.14\textwidth,height=.06\textwidth]{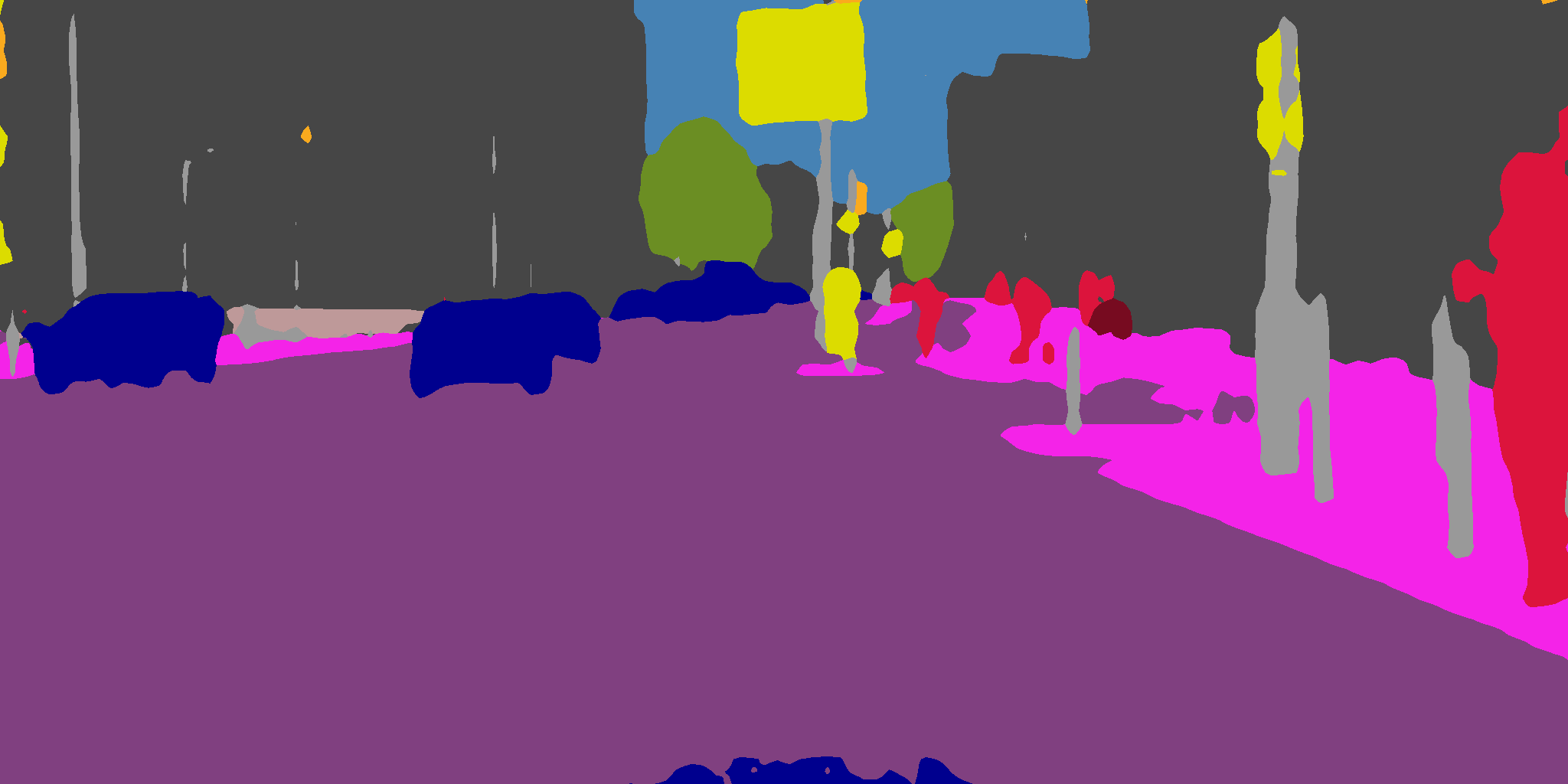} &
			\includegraphics[width=.14\textwidth,height=.06\textwidth]{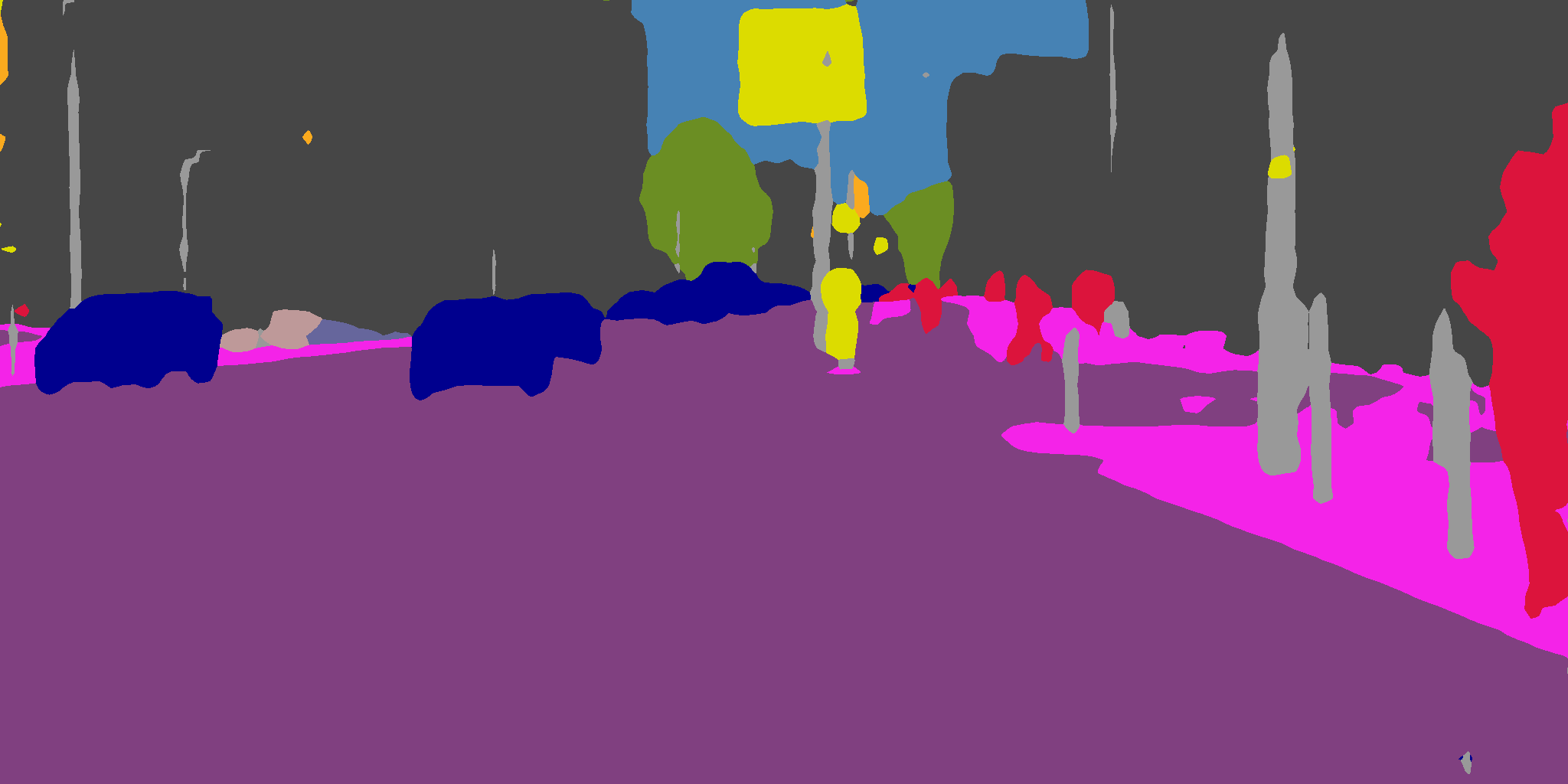} &
			\includegraphics[width=.14\textwidth,height=.06\textwidth]{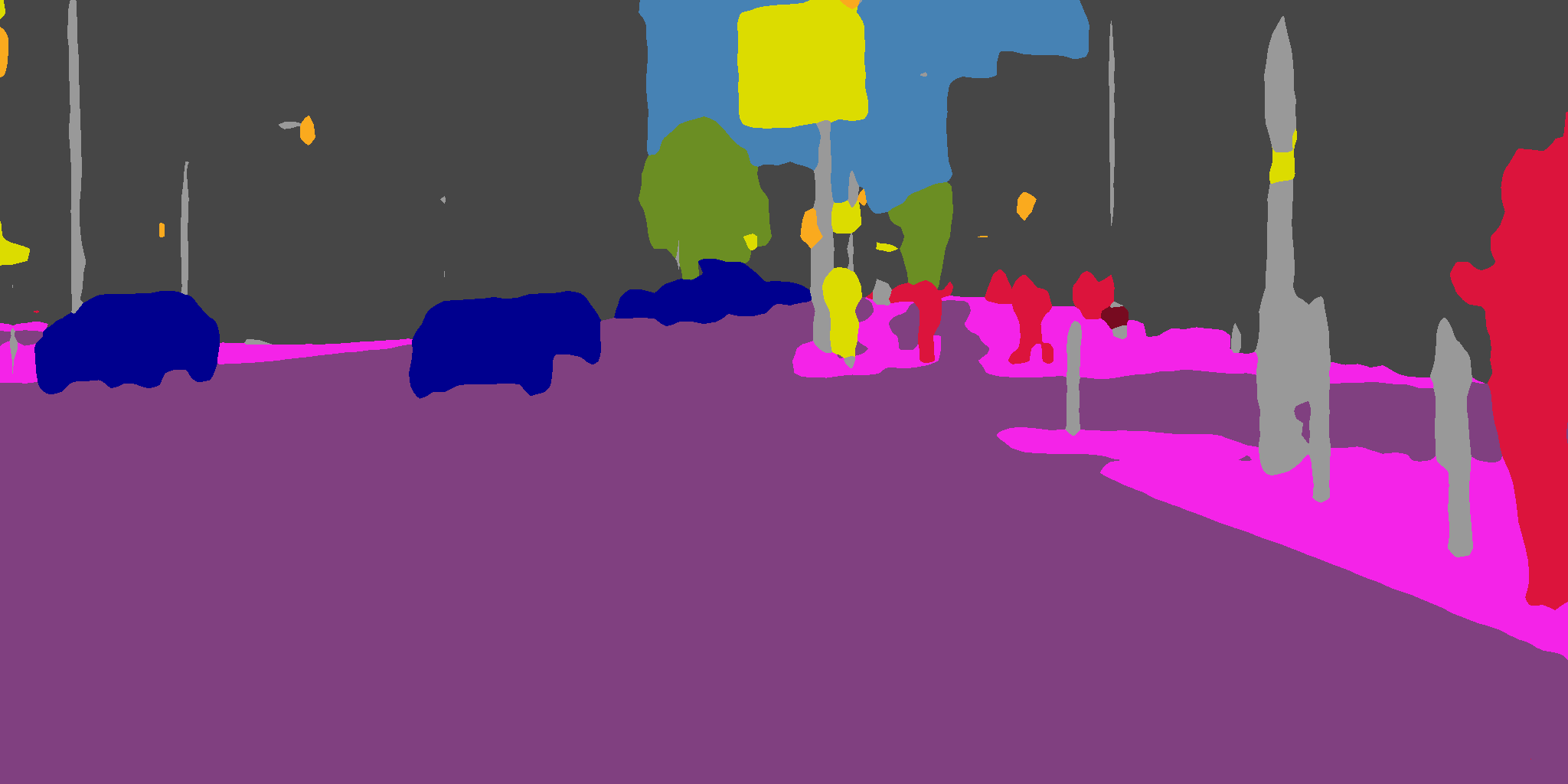} &
			\includegraphics[width=.14\textwidth,height=.06\textwidth]{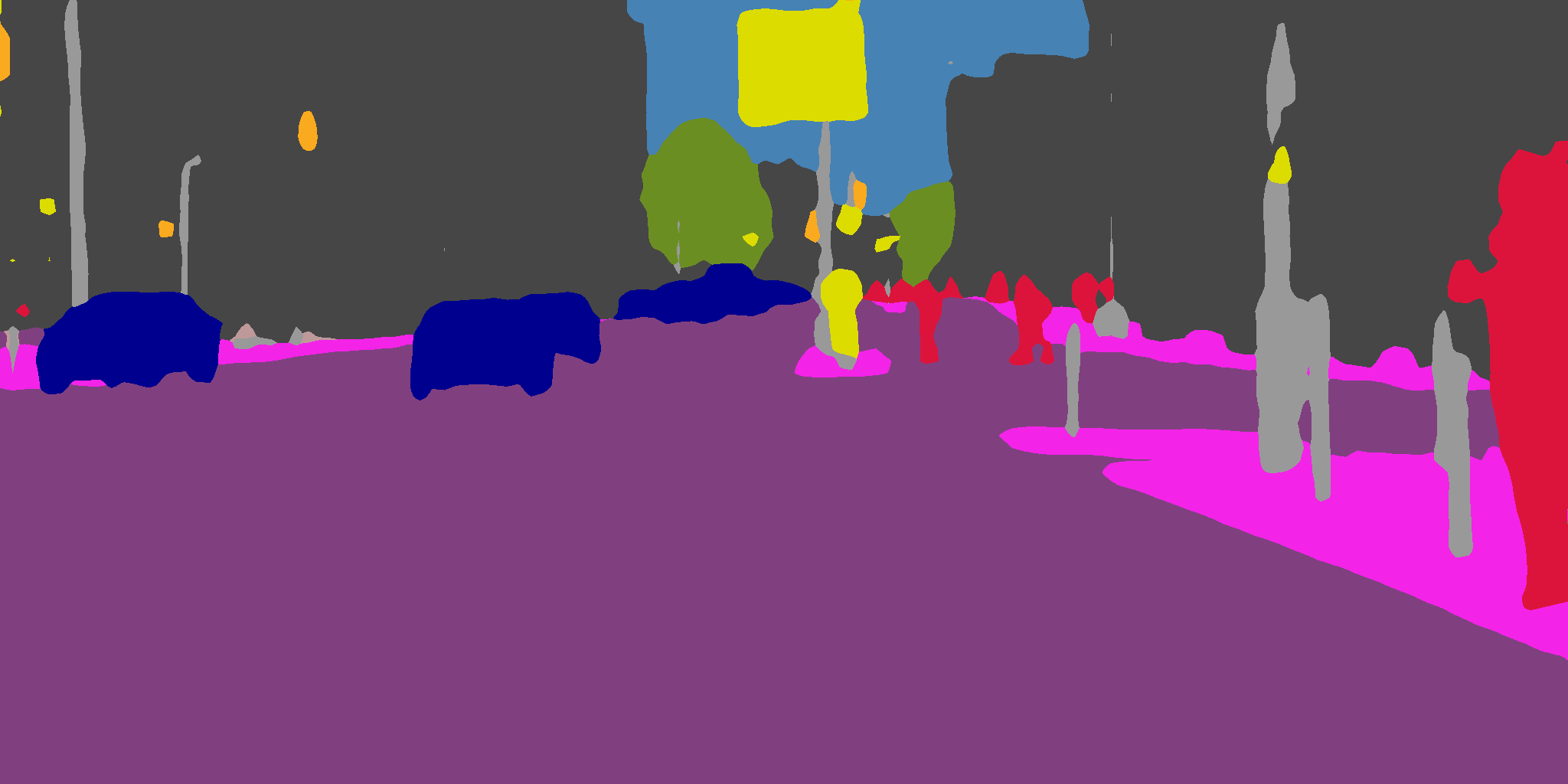}     \\
			\includegraphics[width=.14\textwidth,height=.06\textwidth]{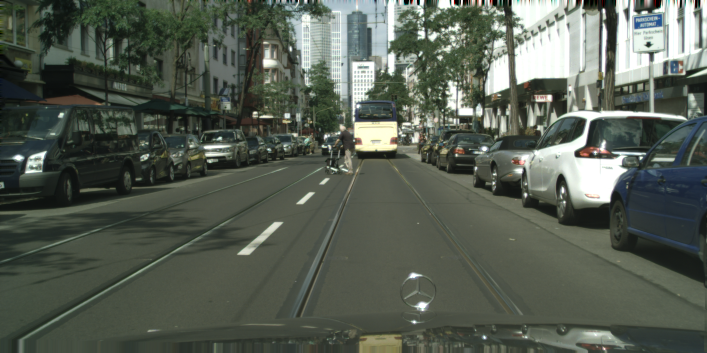} &
			\includegraphics[width=.14\textwidth,height=.06\textwidth]{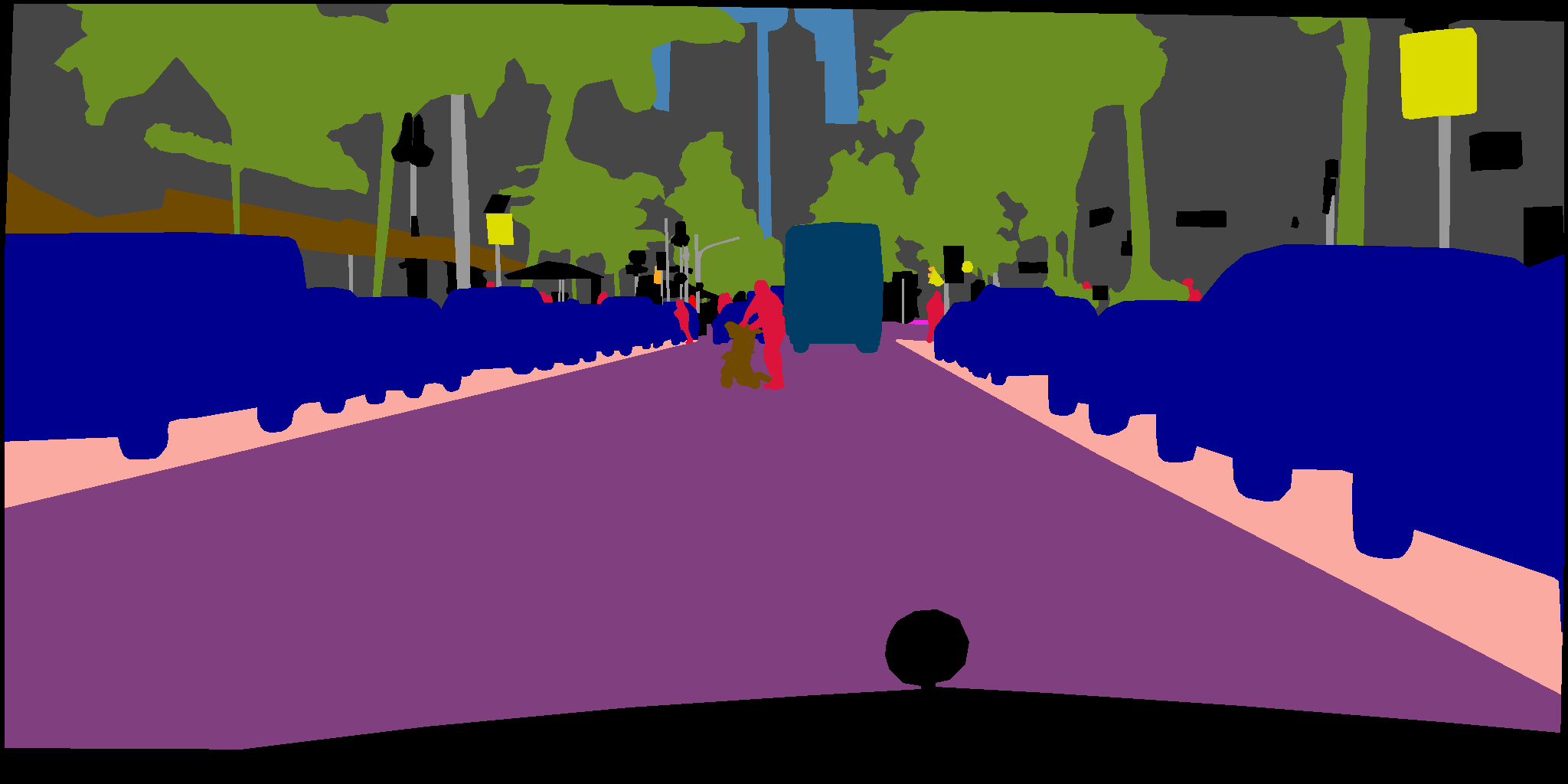} &
			\includegraphics[width=.14\textwidth,height=.06\textwidth]{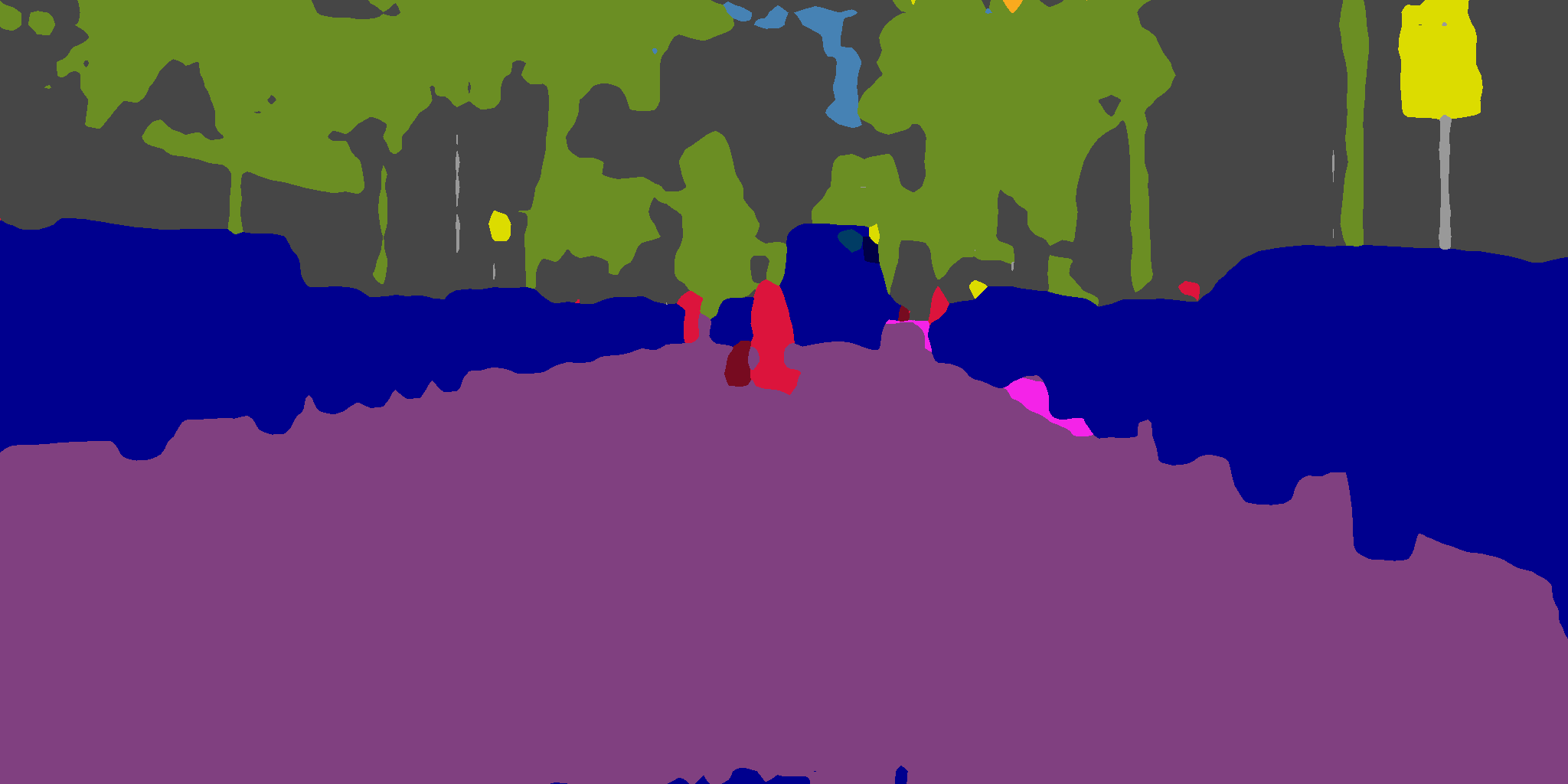} &
			\includegraphics[width=.14\textwidth,height=.06\textwidth]{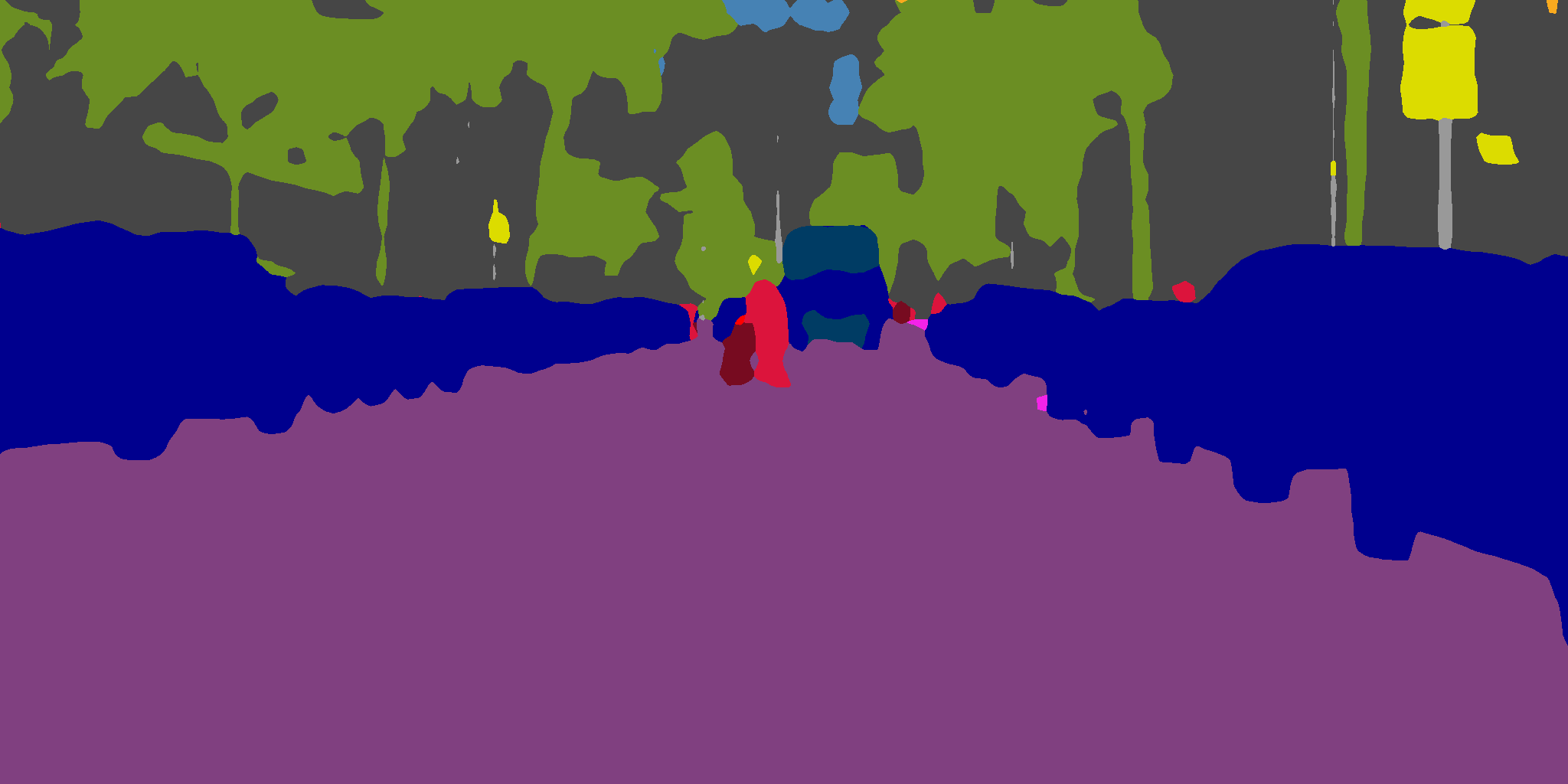} &
			\includegraphics[width=.14\textwidth,height=.06\textwidth]{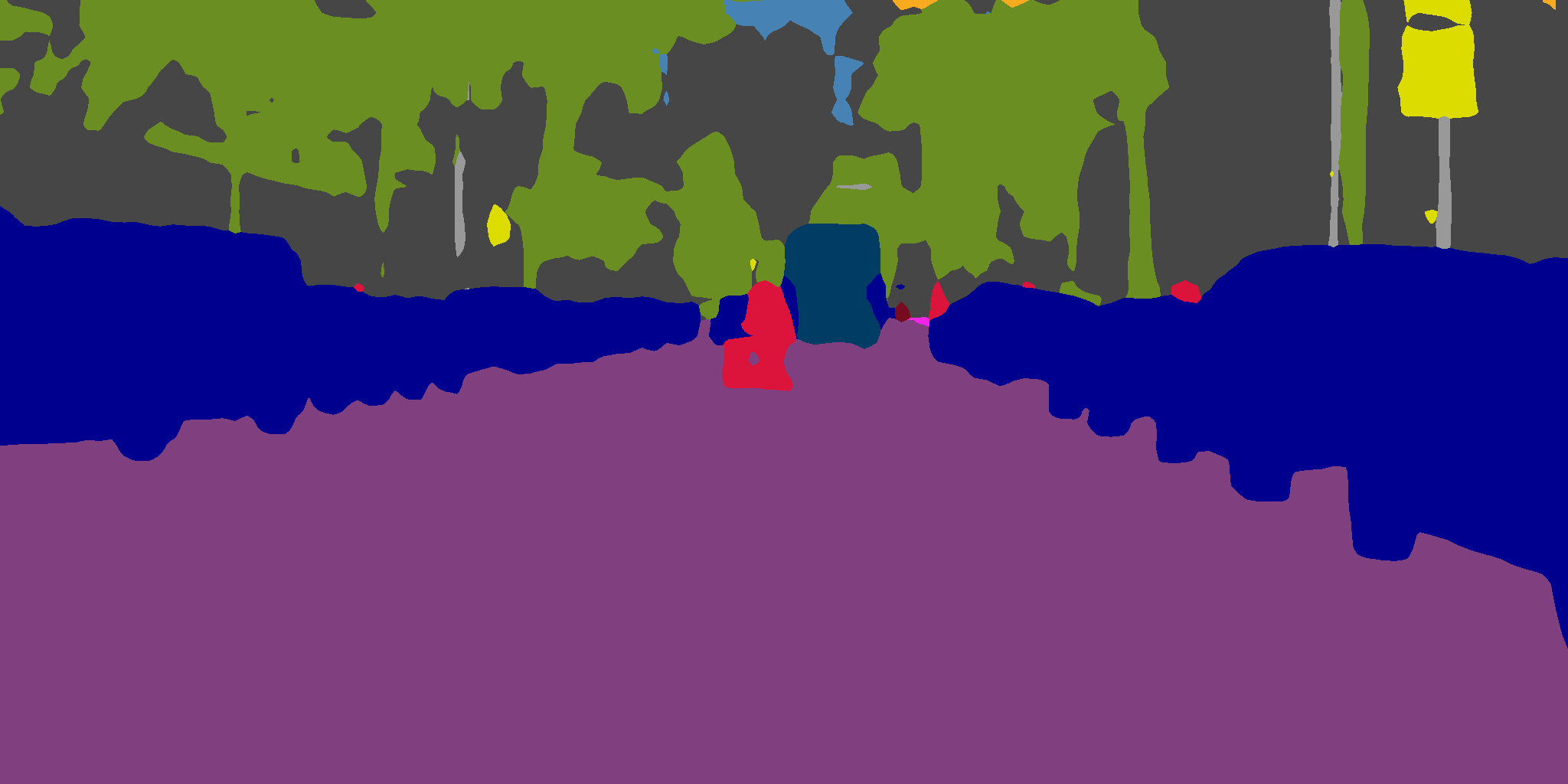} &
			\includegraphics[width=.14\textwidth,height=.06\textwidth]{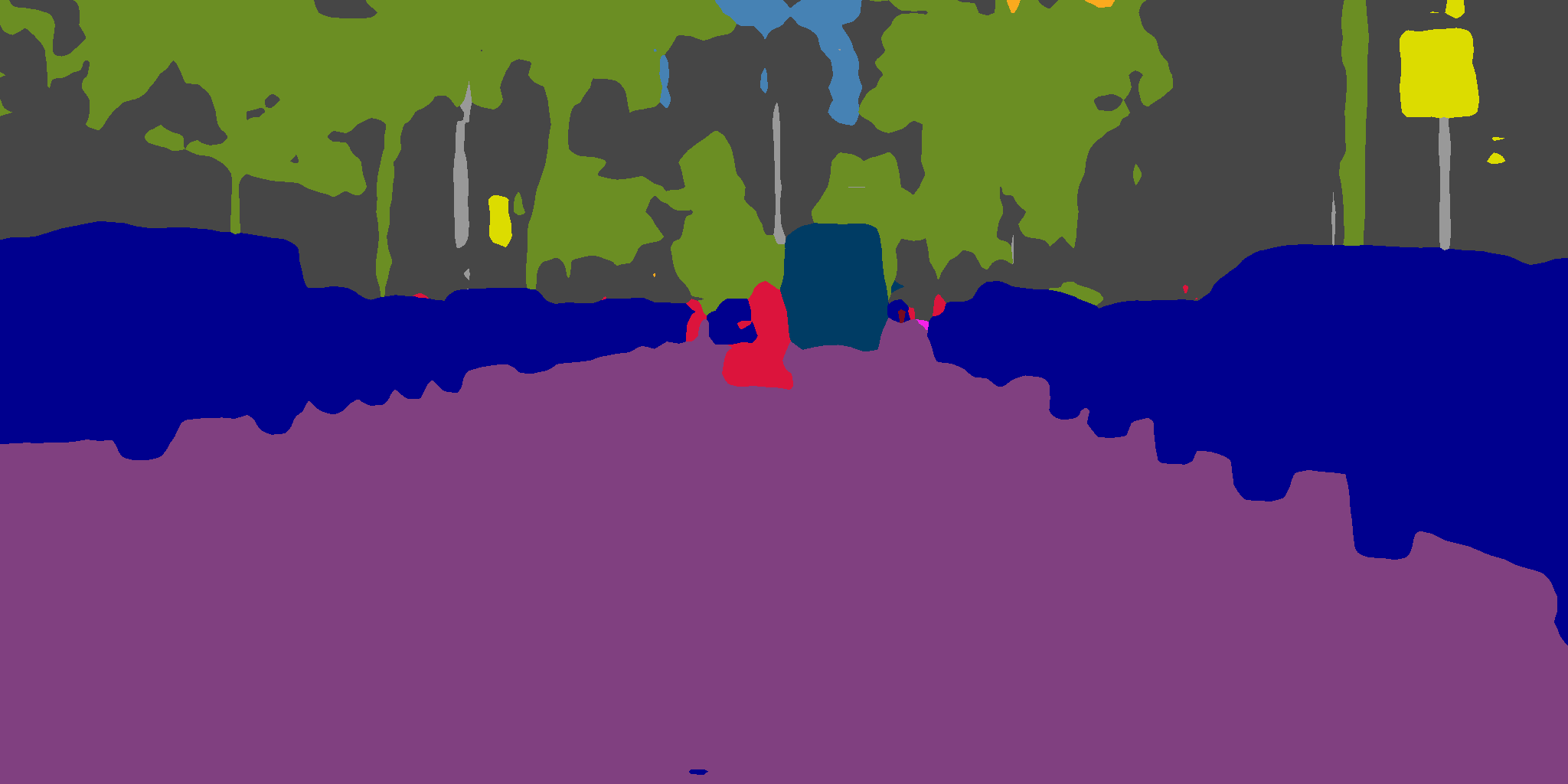}     \\
			image&ground truth&SS-nbt \cite{wang2019lednet}&Model A&Model B&Model C\\
		\end{tabular}
		\caption{Visual resluts on Cityscapes validation set.  More detail results are in supplemental materials.}
		\label{com}
	\end{figure*}
	\subsection{Factorized Convolution Block}
	Designing a factorized convolution block is a popular way to achieve light-weighted segmentation. Techniques like dilated convolution for enlarging receptive field are also important for semantic segmentation models. Our factorized convolution block is inspired by the observation that 1D factorized kernel could be more suitable for spatially less informative features than the spatially informative features. Consider the situation of a $3\times 3$ convolution kernel is replaced by a $3\times 1$ convolution kernel followed by a $1\times 3$ convolution kernel, which could have the same receptive field and fewer parameters. However, neglecting the information lost of crossing the activate function between the two 1D convolution kernel, it could be a rank-1 approximation for the $3\times 3$ convolution kernel. Assuming that different spatial semantic regions have different features, if the dilation of convolution kernel is one or small, the kernel may not lay across multiple different spatial semantic regions and the receptive features are less informative and simple so that the rank-1 approximation is more likely to be effective, and vice versa. 
	
	Therefore, the convolution kernel with large dilation will receive complex or spatially informative long-range features (features separated with a large dilation) in space, and it needs more parameters in space. Meanwhile, a convolution kernel with small dilation will receive simple or less informative short-range features in space, and fewer parameters in space are enough. Our FCB (Fig. \ref{fig:FCB:c}) first deals with short-range and spatially less informative features with 1D factorized convolution in two split groups, which is fully connected in channel, so the factorized convolution reduces the parameter and computation a lot. To enlarge the receptive field, our FCB utilizes 2D kernel with larger dilation and use  depthwise separable convolution to reduce the parameter and computation.  A channel shuffle operation is set at last because there is a residual connection after the point-wise convolution. In total, FCB uses a low-rank approximation (1D kernel) in space for short-range features and depth-wise spatial 2D dilated kernel for long-range features, which lead to  more light-weight, efficient and powerful feature extraction.
	
	Compared with other factorized convolution blocks (Fig. \ref{fig:FCB}), our FCB has a more elaborate design, fewer parameters, less computation, and faster speed, which will be shown in the experiment part further.
	
	\subsection{SVN Module}
	
	A light-weighted model can hardly achieve powerful feature extraction as a big network. Therefore, to produce reduced, robust and representative features and combine them into non-local modules is an essential way to explore the efficient non-local mechanism for light-weighted semantic segmentation. We revisit the non-local mechanism in the form of Query-Key-Value and claim that using the reduced and representative features as the Keys and the Values could reduce computation and memory, as well as maintain effectiveness.
	
	Our SVN module is presented in Fig. \ref{Arc:b}. We reduced the cost in two ways, which are forming a bottleneck by Conv1 and Conv2 to reduced channels for non-local operation and replacing the Keys and Values by their regional dominant singular vectors. The proposed SVN consists of two branches. The lower branch is a residual connection from the input. The upper branch is the bottleneck of our reduced non-local operation. In the bottleneck, we divide the feature maps into spatial sub-regions. We divide $C'\times H\times W$ feature maps into $S=\frac{H\times W}{H'\times W'}$ ($S\ll N=WH$) spatial sub-regions with a scale of $C'\times H'\times W'$ . For each sub-region, we flatten it into  a $C'\times (H'W')$ matrix, then use the Power Iteration Algorithm \ref{PI} to calculate their left dominant singular vectors ($C'\times 1$) efficiently. As is mentioned in Sec 3.1, rotating columns does not affect the left orthogonal matrix, so the left dominant singular vector is agnostic to the way of  flattening and this property is similar to pooling. Then the regional dominant singular vectors are used as the $Keys \in R^{C'\times S}$ and $Values\in R^{C'\times S}$ for the non-local operation, where a smaller S means less computation, and the $Queries\in R^{C'\times N}$ are positional vectors ($C'\times 1$) from the feature maps before dominant singular vector extraction. To enhance the reduced non-local module, we also perform multi-scale region extraction and gather dominant singular vectors from different scales as the Keys and Values (see  Fig. \ref{Arc:b} and Equation \ref{zhizhang}).
	\begin{equation}
	O_i =\sum_{V_j, K_j\in S_1\cup ... S_n}{dot(Q_i,K_j)V_j}
	\label{zhizhang}
	\end{equation}	
	where $\bm{O_i}$ is the output of SVN, $S_n$s are collections of regional dominant singular vectors from their related scales, the regional dominant vectors are used as both the Keys ($\bm{K_j}$) and Values  ($\bm{V_j}$), $\bm{Q_i}$ is a Query from feature maps before dominant singular vectors extraction, and our SVN uses dot product.
	
	As is illustrated above, the SVN module forms a reduced and effective non-local operation by bottleneck structure and reduced and representative regional dominant singular vectors. The regional dominant singular vector could be the most representative for a region of feature maps. Since some works \cite{Asymmetric} utilize pooling as the Keys and Values, we will compare the pooling, single-scale and multi-scale region extraction in our structure in the ablation experiments.
	
	\section{Experiments}
	We conduct experiments to demonstrate the performance of our FCB block and SVN module and the state-of-the-art trade-off among light-weight, accuracy and efficiency of our proposed segmentation architecture. For ablation, we denote our LRNNet without SVN, with single-scale SVN and multi-scale SVN as Model A, Model B and Model C, respectively. 
	\begin{figure}
		\centering
		\subfigure[Non-bt-1D]{\label{fig:FCB:a}\includegraphics[width=.11\textwidth,height=.18\textwidth]{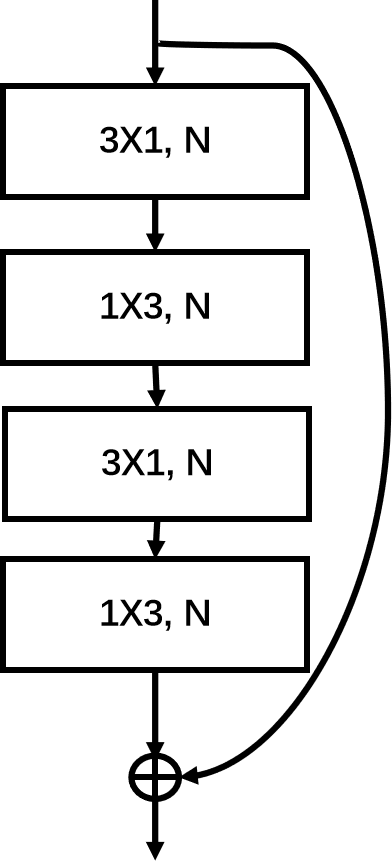}}
		\subfigure[SS-nbt]{\label{fig:FCB:b}\includegraphics[width=.14\textwidth,height=.18\textwidth]{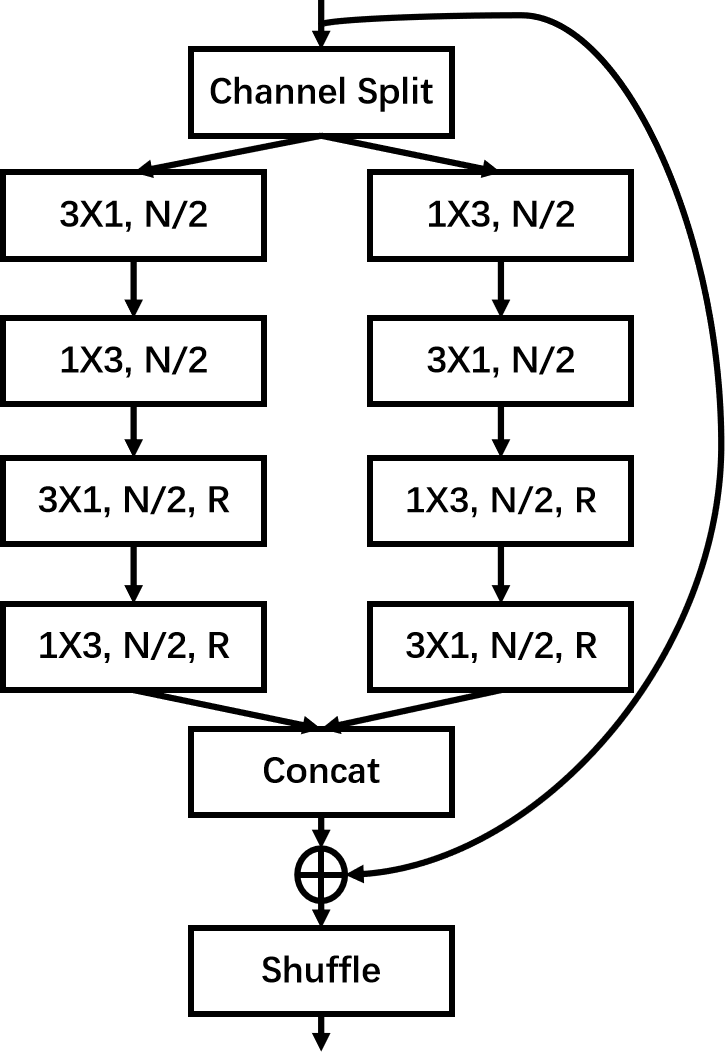}}
		\subfigure[Our FCB]{\label{fig:FCB:c}\includegraphics[width=.16\textwidth,height=.18\textwidth]{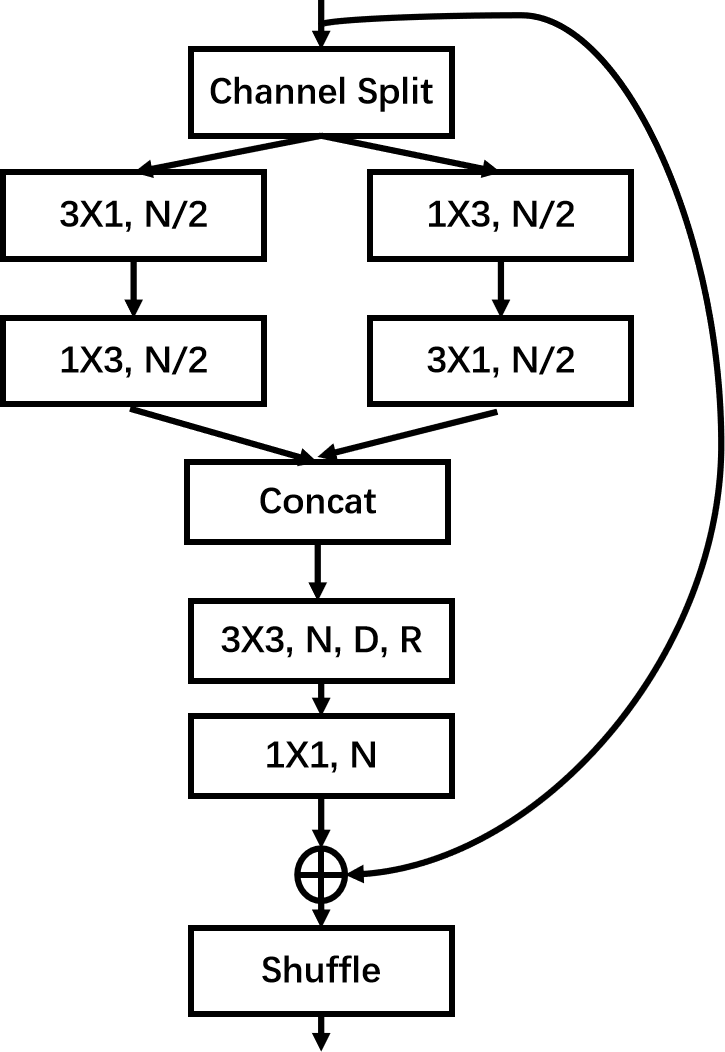}}
		\caption{ Comparison on factorized convolution with other methods. The 'R' is the dilation of kernel and 'D' represents a depth-wise convolution, 'N' is the umber of output channel. (a) Non-bt-1D of ERFNet \cite{ERFNet}. (b) SS-nbt of LEDNet \cite{wang2019lednet}. (c) Our FCB.}
		\label{fig:FCB}
	\end{figure}

	\subsection{Datasets and Settings}
	The Cityscapes dataset \cite{cityscapes} consists of high-quality pixel-level annotations of 5000 street scenes 2048 $ \times $ 1024 images and there are 2975, 500 and 1525 images in the training set, validation set and test set respectively. Following the light-weighted approaches \cite{wang2019lednet,ERFNet}, we adopt 512$\times$1024 subsampled image for testing. The CamVid dataset\cite{Camvid} contains 367 training, 101 validating and 233 testing images with a resolution of 960$\times $720, but we follow the setting as \cite{ENet,Segnet} using  480$\times $360 images for training and testing. 
	
	We implement all our methods using Pytorch \cite{pytorch} on a single GTX 1080Ti. Following \cite{deeplabv3}, we employ a poly learning rate policy and the base learning rate is 0.01. The batch size is set to 8 for all training. For CamVid  testing and evaluation on Cityscapes validation set, we take 250k iterations for training to study our network quickly. And we only train our model on fine annotations for Cityscapes test set with 520K iterations. 
	\begin{table}[t]
		\begin{center}
			\caption{Evaluation on Cityscapes validation set, including accuracy, inference time, parameter size and computation.}
			\label{sim}
			\begin{tabular}{|c|c|c|c|c|}
				\hline
				Model & mIoU & Times(ms)&Para(M)&GFLOPS\\
				\hline
				SS-nbt \cite{wang2019lednet} & 69.6 & 14&0.95&11.7\\
				Model A & 70.6 & 13&0.67&8.48\\
				Max Pooling & 70.2 & 14&0.68&8.54\\
				Avg Pooling & 70.3 & 14&0.68&8.54\\
				Model B & 71.1 & 14&0.68&8.57\\
				Model C & 71.4 & 14&0.68&8.58\\
				\hline
			\end{tabular}
		\end{center}
		
	\end{table}
	\begin{table}[t]
		\begin{center}
			\caption{Ablation for sub-region selection of SVN on Cityscapes val set comparing with standard non-local. SS means single-scale and MS means Multi-scale.}
			\label{select}
			\begin{tabular}{|c|c|c|c|c|}
				\hline
				Sub-region & mIoU & Times(ms)&GFLOPS\\
				\hline
			    non-local (64$\times$128)& 71.2 & 22&12.5\\
				SS (16$\times$16) & 71.1 & 15&8.68\\
				SS (8$\times$8) & 71.1 & 14&8.57\\
				SS (4$\times$4) & 70.8 & 14&8.51\\
				MS (8$\times$8+4$\times$4) & 71.4 & 14&8.58\\				MS (8$\times$8+4$\times$4+2$\times$2) & 71.4 & 14&8.59\\
				\hline
			\end{tabular}
		\end{center}
		
	\end{table}	
	\subsection{Ablation Study for FCB}
	Comparing with other factorized convolution blocks shown in Figure \ref{fig:FCB}, ERFNet \cite{ERFNet} and LEDNet \cite{wang2019lednet} simply use 1D factorized kernel to deal with short-range and long-range (with dilation) features. As is analyzed in Section 3.3, our FCB deals with short-range features with 1D factorized kernel and long-range features with the 2D depth-wise kernel. We compare our FCB (Model A) with SS-nbt from LEDNet \cite{wang2019lednet} in the same  architecture. As shown in Table \ref{sim} and \ref{Camvid}, our FCB (Model A) achieves better accuracy with lower parameter size, computation and inference time comparing with SS-nbt which using 1D factorized kernel for both short-range and long-range features. Visual examples are in Fig. \ref{com}.
	\subsection{Ablation Study for SVN}
	Table \ref{select} shows the performances of different sub-region choices of our SVN module and the standard non-local. Balancing accuracy, speed and computation cost, we choose 64 (8$\times8$) sub-region as single-scale SVN (Model B) and 8$\times$8+4$\times$4 sub-regions for multi-scale SVN (Model C).
	
	We analyze the efficiency of our SVN. Since Algorithm \ref{PI} converges efficiently, we set the T as 2, whose computation complexity is $O(C(WH)T)$. The features in our bottleneck on Cityscapes is $32\times 64 \times 128$, and the computation is 4.0 GFLOPS in standard non-local operation neglecting the convolution and the complexity is $O(C(WH)^{2})$. For our reduced non-local operation, the complexity is $O(C(WH)S)$, where S ($S\ll WH$) is the number of the Keys and Values. For single-scale SVN (Model B) and pooling, we divide feature maps into 64 sub-regions and the computation is 32 MFLOPS. For multi-scale SVN, feature maps into 64  and 16 sub-regions, and the computation is 40 MFLOPS. The cost of Power Iteration in single-scale and multi-scale SVN are 1 MFLOPS and 2 MFLOPS, respectively.
	\begin{table}[t]
		\begin{center}
			\caption{Evaluation on Cityscapes test set.}
			\label{City}
			\begin{tabular}{|c|c|c|c|c|c|c|}
				\hline \multirow{1}[10]{1cm}
				{Model} & \rotatebox{-90}{Subsample}& \rotatebox{-90}{Pre-trained}& \rotatebox{-90}{\bf{mIoU}} & \rotatebox{-90}{FPS}&\rotatebox{-90}{Para(M)}&\rotatebox{-90}{GFLOPS}\\
				\hline
				SegNet \cite{Segnet} & 3 & N&57.0&16.7&29.5&286\\
				ENet \cite{ENet} & 3 & N&58.3&\bf{135}&\bf{0.37}&3.8\\
				FRRN \cite{FRRN} & 2 & N&71.8&0.25&24.8&235\\
				ICNet \cite{ICNet} & 1 & Y&69.5&30.3&26.5&28.3\\
				ERFNet \cite{ERFNet} & 2 & N&68.0&41.7&2.1&21.0\\
				CGNet \cite{CGNet} & 3 & Y&64.8&50.0&0.5&6.0\\
				BiSeNet \cite{BiSeNet} & 4/3 & N&68.4&-&5.8&14.8\\
				DFANet \cite{li2019dfanet} & 2 & Y&70.3&-&7.8&\bf{1.7}\\
				LEDNet \cite{wang2019lednet} & 2 & N&69.2&71&0.94&11.5\\
				\hline
				Model A  &2& N&70.6&76.5&0.67&8.48\\
				Model B  &2& N&71.6&71&0.68&8.57\\
				Model C  &2& N&\bf{72.2}&71&0.68&8.58\\
				\hline
			\end{tabular}
		\end{center}
		
	\end{table}
	
	We compare using regional dominant singular vectors (Model B) with using pooling features in single scale ($8\times 8$) (Table \ref{sim}) to show the effectiveness of dominant singular vectors. Results on Cityscapes validation are shown in  Table \ref{sim}. Comparison of Model single-scale SVN (Model B) and pooling (max or average) shows that regional singular vectors are effective for our network with a light-weighted encoder with 0.5 mIoU improvement and additional 0.09 GFLOPS, while using pooling can not provide representative features for a light-weighted network. And the multi-scale SVN (Model C) further improves the result to 71.4\% mIoU with a little cost on inference time and computation.
	\subsection{Comparison with Other  Methods}
	We compare our LRNNet with other light-weighted methods on Cityscapes and Camvid test sets in terms of parameter size, accuracy, speed and computation. We only report the speed of methods on the open-source deep-learning framework, such as Pytorch, TensorFlow and Caffe, because they have comparable implemented performance, but have a large gap comparing with the non open-source deep-learning framework of those works \cite{li2019dfanet,BiSeNet} and details are in supplemental material. Results are shown in Table \ref{City} and \ref{Camvid}. ”-” indicates that the speed is not achieved by open-source deep learning frameworks or not provided. Our network constructed by FCB (Model A) achieves 70.6\% mIoU and 76.5 FPS on Cityscapes test set with only 0.67M parameters and 8.48 GFLOPS, which is more light-weighted and efficient than ERFNet \cite{ERFNet} and LEDNet \cite{wang2019lednet} with better accuracy. With single-scale (Model B) and  multi-scale SVN (Model C), LRNNet achieves 71.6\% and 72.2\% mIoU on Cityscapes test set with a little cost on speed and efficiency, respectively. Our LRNNet with multi-scale SVN achieves 69.2\% mIoU with only 0.68M parameters on CamVid test set. All results show the state-of-the-art trade-off among parameter size, speed, computation and accuracy of our LRNNet. Visual comparison can be viewed in Fig. \ref{fig1}.

	\begin{table}[!htb]
		\begin{center}
			\caption{Evaluation on Camvid test set.}
			\label{Camvid}
			
			\begin{tabular}{|c|c|c|c|c|c|}
				\hline \multirow{1}[10]{1cm}
				{Model} & \rotatebox{-90}{Input Size}& \rotatebox{-90}{Pre-trained}& \rotatebox{-90}{mIoU} & \rotatebox{-90}{FPS}&\rotatebox{-90}{Para(M)}\\
				\hline
				SegNet \cite{Segnet} & $360\times 480$ & N&46.4&46&29.5\\
				ENet \cite{ENet} & $360\times 480$ & N&51.3&-&\bf{0.37}\\
				ICNet \cite{ICNet} & $720\times 960$ & Y&67.1&27.8&26.5\\
				CGNet \cite{CGNet} & $360\times 480$ & Y&65.6&-& 0.5\\
				BiSeNet \cite{BiSeNet} &$720\times 960$ & N&65.6&-&5.8\\
				DFANet \cite{li2019dfanet} & $720\times 960$ & Y&64.7&-&7.8\\
				\hline
				SS-nbt \cite{wang2019lednet}& $360\times 480$  & N&66.6&\bf{77}&0.95\\
				\hline
				Model A  & $360\times 480$  & N&67.6&\bf{83}&0.67\\
				Model B  & $360\times 480$  & N&68.9&77&0.68\\
				Model C  & $360\times 480$  & N&\bf{69.2}&76.5&0.68\\
				\hline
			\end{tabular}
		\end{center}
		
	\end{table}
	
	\section{Conclusion}
	We have proposed LRNNet for real-time semantic segmentation. The proposed FCB unit explores a proper form of factorized convolution block to deal with short-range and long-range features, which provides light-weighted, efficient and powerful feature extraction for the encoder of our LRNNet. Our SVN module utilizes regional dominant singular vectors to construct the efficient reduced non-local operation, which enhances the decoder with a very low cost. Extensive experimental results have validated our state-of-the-art trade-off in terms of parameter size, speed, computation and accuracy.
	%
	\section{ACKNOWLEDGEMENT}
	This paper is supported by NSFC (No.61772330, 61533012, 61876109), the pre-research project (No.61403120201), Shanghai Key Laboratory of Crime Scene Evidence (2017XCWZK01) and the Interdisciplinary Program of Shanghai Jiao Tong University (YG2019QNA09). 
	\scriptsize
	\bibliographystyle{IEEEbib}
	\bibliography{icme2020template}
	
\end{document}